\documentclass{article}

\usepackage[nonatbib, final]{neurips_2022}
\usepackage{float}

\usepackage[
backend=biber,
]{biblatex}
\usepackage{xfrac}

\PassOptionsToPackage{showframe}{geometry}

\usepackage[utf8]{inputenc} 
\usepackage[T1]{fontenc}    
\usepackage{hyperref}       
\usepackage{url}            
\usepackage{booktabs}       
\usepackage{amsfonts}       
\usepackage{nicefrac}       
\usepackage{microtype}      
\usepackage{xcolor}         
\usepackage{graphicx} 
\usepackage{caption,subcaption}
\usepackage[showframe]{geometry}
\usepackage{booktabs}
\usepackage{tabulary,lipsum}
\usepackage[nameinlink]{cleveref}

\title{Tracking the Dynamics of the Tear Film Lipid Layer}

\author{%
  Tejasvi Kothapalli\textnormal{$^{1,2}$}
   \And
   Charlie Shou\textnormal{$^{1}$}
   \And
   Jennifer Ding\textnormal{$^{1}$}
   \And
   Jiayun Wang\textnormal{$^{1,2}$}
   \And
   Andrew D. Graham\textnormal{$^{1}$}
   \And
   Tatyana Svitova\textnormal{$^{1}$}
   \And
   Stella X. Yu\textnormal{$^{1,2,3}$}
   \And
   Meng C. Lin\textnormal{$^{1}$}
   \AND
   \\\textnormal{$^{1}$UC Berkeley, $^{2}$ICSI, $^{3}$University of Michigan} \\
   \texttt{\{tejasvi.kothapalli, charlie\_shou, jennifer.ding, peterwg,}\\
   \texttt{agraham, svitova, stellayu, mlin\}@berkeley.edu}
}

\begin{document}

\maketitle

\begin{abstract}
  Dry Eye Disease (DED) is one of the most common ocular diseases: over five percent of US adults suffer from DED \cite{farrand2017prevalence}. Tear film instability is a known factor for DED, and is thought to be regulated in large part by the thin lipid layer that covers and stabilizes the tear film. In order to aid eye related disease diagnosis, this work proposes a novel paradigm in using computer vision techniques to numerically analyze the tear film lipid layer (TFLL) spread. Eleven videos of the tear film lipid layer spread are collected with a micro-interferometer and a subset are annotated. A tracking algorithm relying on various pillar computer vision techniques is developed. Our method can be found at \href{https://easytear-dev.github.io/}{https://easytear-dev.github.io/}.
\end{abstract}

\section{Introduction}
 
The fast and uniform spreading of tear film over the ocular surface and tear film stability
are of vital importance for acute vision and overall ocular health. DED is one
of the most frequently encountered ocular morbidities. According to the TFOS DEWS II
definition: \textit{“Dry eye (DE) is a multifactorial disease of the ocular surface characterized by a loss of homeostasis of the tear film, and accompanied by ocular symptoms, in which tear film instability and hyperosmolarity, ocular surface inflammation and damage, and neurosensory abnormalities play etiological roles.”} \cite{craig2017tfos} There are many factors involved in the etiology of DED.

Tear film instability is perhaps the most widely studied factor, and depends on the interplay
of eyelid biomechanical action and the interfacial properties of the intricate multi-layered tear
film spread over the ocular surface. Complex mixtures of oily substances known as tear lipids
form the outermost layers, measuring approximately 30-100 nm in thickness on the 1-10 µm
human tear film.

A major impediment in modern DED diagnostics is the absence of numerical tools to clearly
differentiate dry eye subtypes, i.e., evaporative, aqueous-deficient, mixed, and DE caused by
excessive drainage. Micro-interferometric devices permit visualization and recording of the tear
lipid layer and its dynamics during inter-blink periods. The lipid layer is a crucial factor in tear
film stability, and thus may hold clues to the diagnosis of this globally impactful disease.

Numerical evaluation of vertical and horizontal propagation rates of tear
lipid films during inter-blinks can provide valuable information regarding tear lipid film
quality, visco-elasticity, mobility, tear film stability, and may lead to improved DED
diagnosis. We investigate computer vision techniques as suitable instruments for the development
of a novel DED diagnostics paradigm.

\cite{yokoi2008rheology} is the only prior work which studies the tear film lipid layer spread. Their work finds that the height displacement of the film spread can be fit with an exponential decay curve. Additionally, they find that the velocity of the lipid layer spread decreases in proportion to the decrease of tear volume. They collect interference images from the tear film lipid layer with a video-interferometer. The spreading lipid layer is then tracked manually with photo editing software. Our work aims to use modern computer vision to track the spreading lipid layer automatically.

\begin{figure}[H] 
    \centering 
\begin{subfigure}{0.25\textwidth}
  \includegraphics[width=\linewidth]{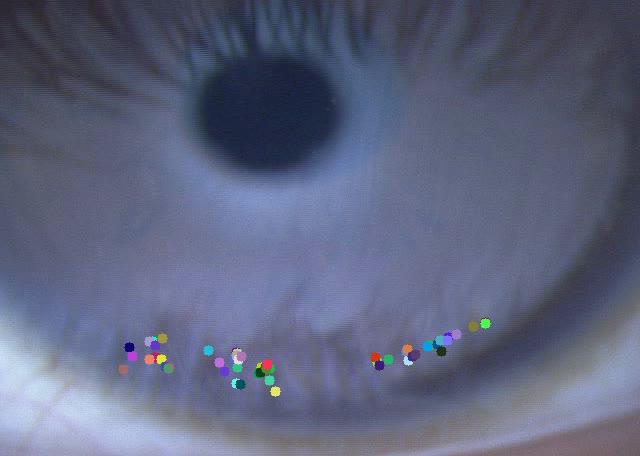}
  \caption{frame 1}
  \label{fig:1}
\end{subfigure}\hfil 
\begin{subfigure}{0.25\textwidth}
  \includegraphics[width=\linewidth]{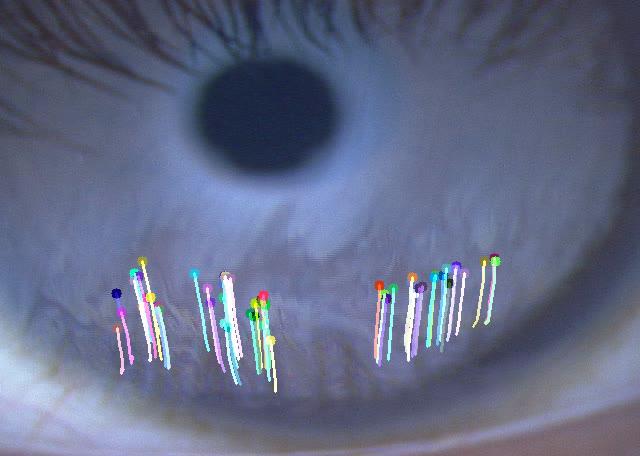}
  \caption{frame 7}
  \label{fig:2}
\end{subfigure}\hfil 
\begin{subfigure}{0.25\textwidth}
  \includegraphics[width=\linewidth]{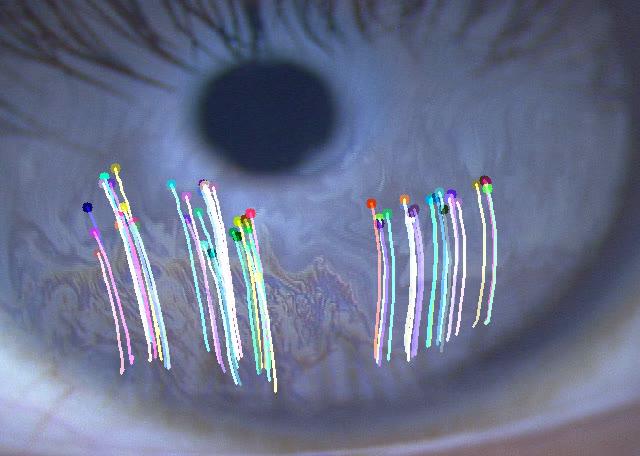}
  \caption{frame 20}
  \label{fig:3}
\end{subfigure}\hfil 
\begin{subfigure}{0.25\textwidth}
  \includegraphics[width=\linewidth]{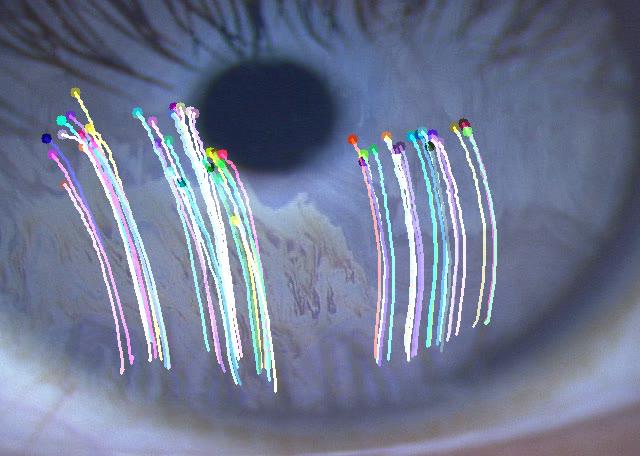}
  \caption{frame 45}
  \label{fig:4}
\end{subfigure}

\medskip
\begin{subfigure}{0.25\textwidth}
  \includegraphics[width=\linewidth]{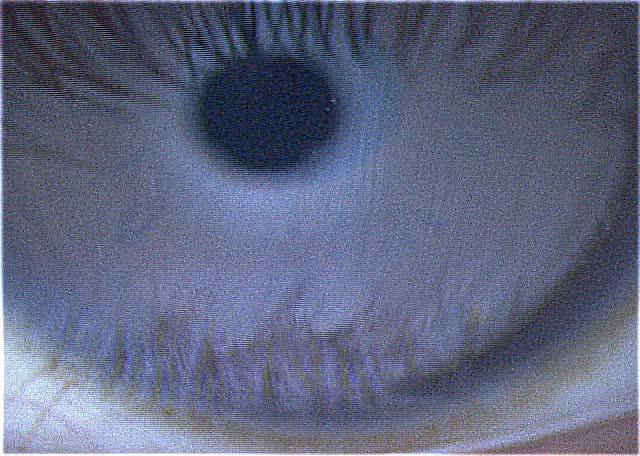}
  \caption{0.03 seconds}
  \label{fig:5}
\end{subfigure}\hfil 
\begin{subfigure}{0.25\textwidth}
  \includegraphics[width=\linewidth]{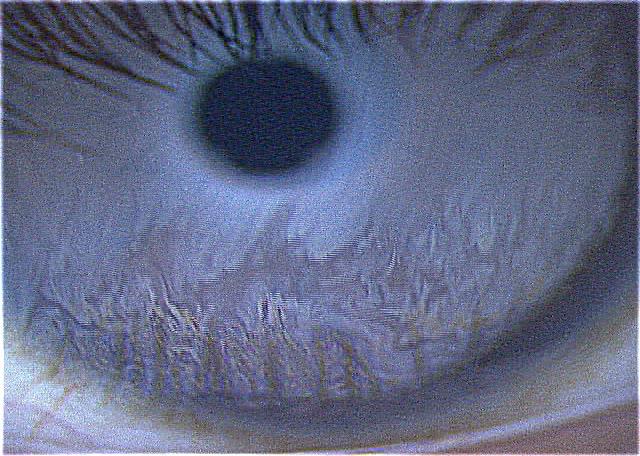}
  \caption{0.23 seconds}
  \label{fig:6}
\end{subfigure}\hfil 
\begin{subfigure}{0.25\textwidth}
  \includegraphics[width=\linewidth]{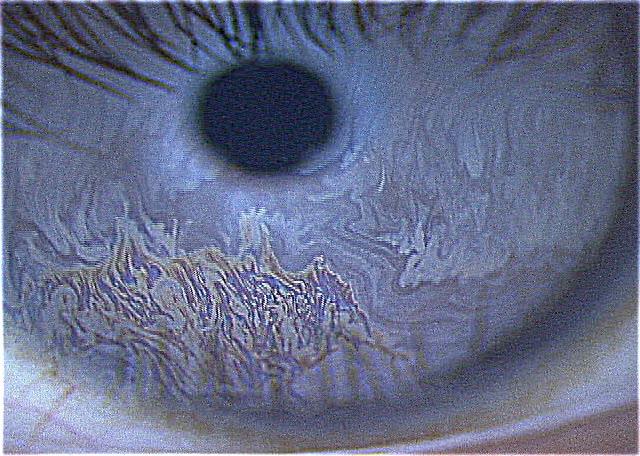}
  \caption{0.66 seconds}
  \label{fig:7}
\end{subfigure}\hfil 
\begin{subfigure}{0.25\textwidth}
  \includegraphics[width=\linewidth]{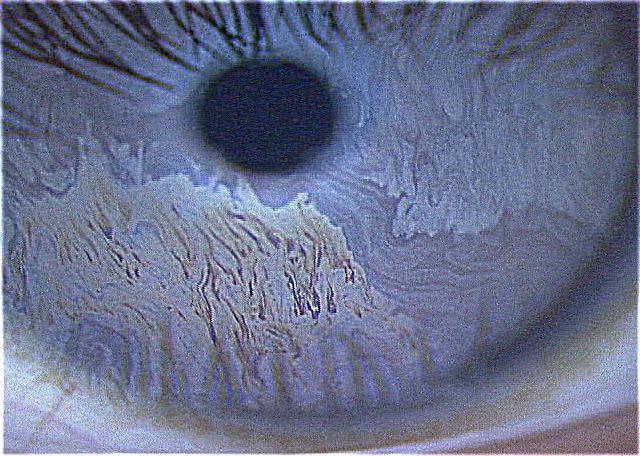}
  \caption{1.5 seconds}
  \label{fig:8}

\end{subfigure}

\caption{\Crefrange{fig:1}{fig:4} show feature points tracking the upward spread of the lipid layer through time. \Crefrange{fig:5}{fig:8} are the same frames except they are sharpened to make the lipid layer more visible. \Crefrange{fig:1}{fig:4} are labeled with the frame number while the corresponding figures in \Crefrange{fig:5}{fig:8} are labeled with the time equivalent. For example, frame 1 in \Cref{fig:1} is the same frame shown in \Cref{fig:5} marked at 0.03 seconds.}
\label{fig:images}
\end{figure}

\section{Methodology}
\subsection{Data collection}\label{data}
The EasyTearView \cite{easytearwebsite} instrument uses multi-source lighting to illuminate patients' tear films for video recording. Videos are collected one eye at a time. The video recording begins with the patient blinking normally and then refraining from blinking for as long as possible. Due to the variability, patients are asked to repeat this process 3 times. The periods where the patient refrains from blinking are defined as inter-blink periods. Each inter-blink period is truncated at 5 seconds when analyzing the videos. During the inter-blink period, the tear film will initially rapidly restructure and spread but it will slowly stagnate in motion and eventually settle in place until the next blink. This rapid spread can be observed in \Cref{fig:images}. It is important to note that the visibility of the tear film lipid layer for each video is variable due to factors such as camera focus, eye color, and patient lipid layer health. 

From the available data, we identify which inter-blink recordings are in focus and do not suffer from artifacts like excessive eye movement. Of the selected inter-blink periods, 5 are randomly picked and the spread of the tear film is annotated. Distinctive edges or corners are picked to be tracked through the inter-blink period. These points are also picked such that they are spread out horizontally across the iris, they start at the bottom of the iris, and are visible until the end of the inter-blink period. Note, that not every frame was explicitly annotated. Instead roughly every tenth frame was annotated, and the rest of the frames' annotations were interpolated with the frames with explicit annotations.

\subsection{Preprocessing}
\label{Preprocessing}

\subparagraph{Blink detection} 
The goal of blink detection is to find the frames where the eyelid is partially or completely occluding the eye. We observe that the video of the eye looks relatively constant aside from when the blinks appear. Thus, we compute the average frame of the whole video and determine the distance of each frame to the average frame. Frames which deviate beyond a threshold are marked as blink frames. With all the blink frames identified, the three inter-blink periods naturally follow.


\paragraph{Image frame alignment}
In order to address the movement of the eye in the video, the frames are aligned within their own respective inter-blink period. For each frame, the pupil center is located using the method described in \cite{kowalski2021hybrid}. All the frames' pupil centers are aligned with the pupil center of the first frame belonging to inter-blink period.

\paragraph{Tear film image enhancement}
In order to enhance the appearance of the lipid layer, four different image processing techniques are individually applied to the image frames. The first method is to \textit{subtract the average frame}. The average frame of the inter-blink period is computed and subtracted from all the frames belonging to that inter-blink period. The second method is to \textit{sharpen} the image by convolving each frame with the unsharp mask filter. \Crefrange{fig:1}{fig:4} shows what the original images look like while \Crefrange{fig:5}{fig:8} show the sharpened version. The third method is to compute the \textit{Laplacian pyramid} and take the second image from the stack in order to only display the frequencies at which the lipid layer is most visible. The fourth method is to perform \textit{local histogram equalization} for each frame in order to increase the contrast of the image.

\subsection{Tracking the tear film lipid layer spread}

\paragraph{Iris segmentation} The lipid layer spread should only be tracked over the iris region where it is most visible. The sclera, pupil, and eyelashes are all regions that need to be avoided when tracking. The bottom edge of the iris is identified with the Active Contour Model \cite{kass1988snakes} which fits a spline to edges and lines. The circular pupil edge is already determined for each frame during image alignment as mentioned in \Cref{Preprocessing}. Finally, a horizontal line is drawn at the top of the found pupil region to demarcate everything above as the eyelash region.

\paragraph{Initial feature points}
With the iris region segmented, feature points on the lipid layer are picked to track across the inter-blink. These feature points are picked on the last frame of the inter-blink period and tracked backwards through the frames. Strong corner feature points are found using FAST (Features from Accelerated Segment Test) \cite{rosten2008faster}.
\paragraph{Optical flow for feature point tracking}

Two different optical flow algorithms are used to track the feature points backwards through the frames. The first algorithm is the Lucas-Kanade method \cite{lucas1981iterative} which computes the optical flow for a sparse set of feature points. The second algorithm is Farneback's algorithm \cite{farneback2003two} which is a dense optical flow method that computes the flow for all the points in the frame. Given that we still want to track a sparse set of feature points even when computing dense optical flow, the corresponding points are indexed from the dense optical flow estimation map for each frame and updated accordingly.

The two different optical flow algorithms are individually applied on the four enhanced videos along with the original video to yield a total of ten different sets of tracked points. Each set of tracked points is filtered such that the points begin near the bottom of the iris as seen in \Cref{fig:1}. Furthermore, only the highest points in the last frame are considered and everything below these points are disregarded as seen in \Cref{fig:4}. This way the full motion of the tear film is analyzed.
\begin{figure}[H]
    \centering 
    \begin{subfigure}{0.49\textwidth}
      \includegraphics[width=\linewidth]{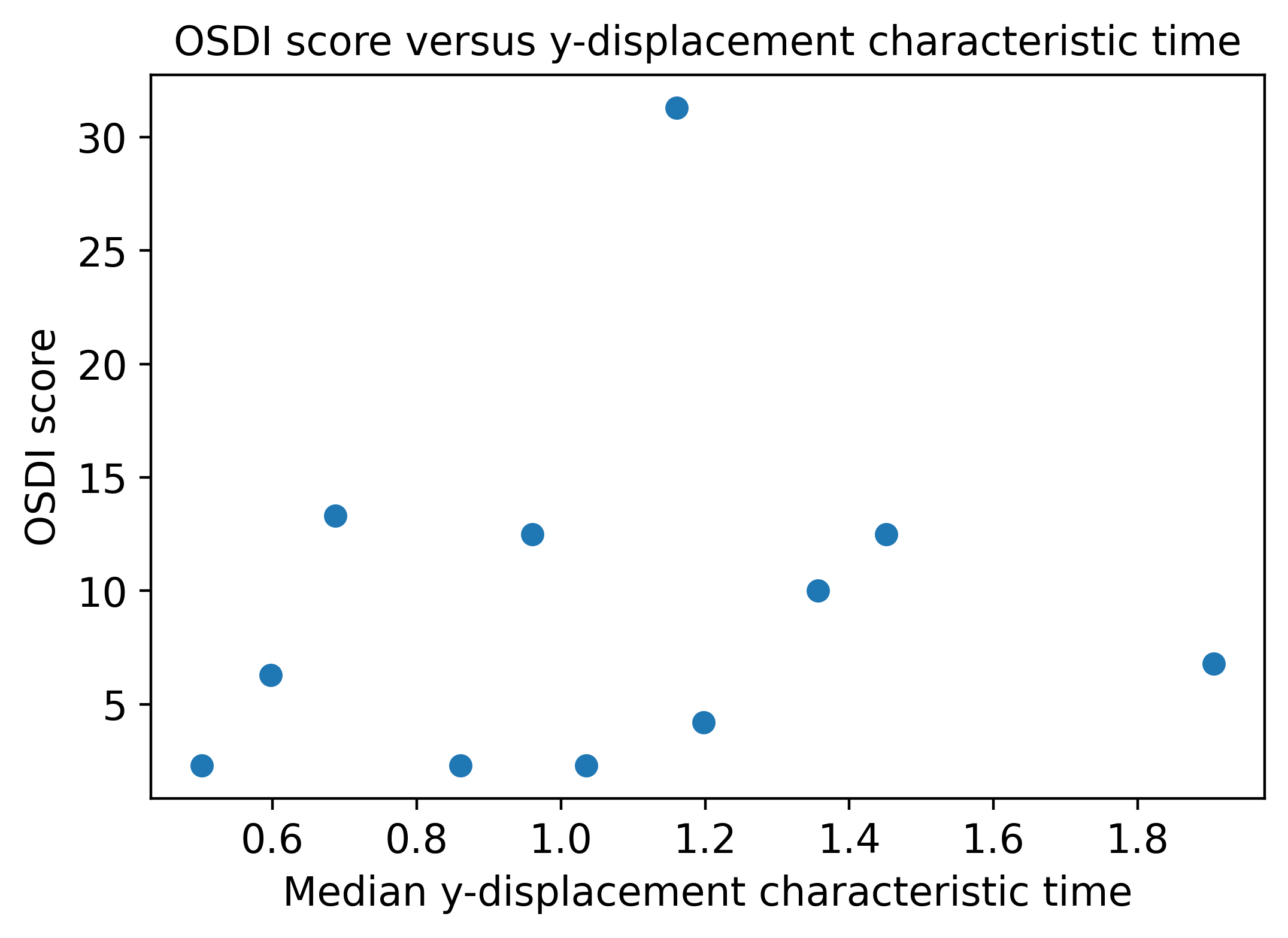}
      \caption{}
      \label{fig:9}
    \end{subfigure}\hfil 
    \begin{subfigure}{0.49\textwidth}
      \includegraphics[width=\linewidth]{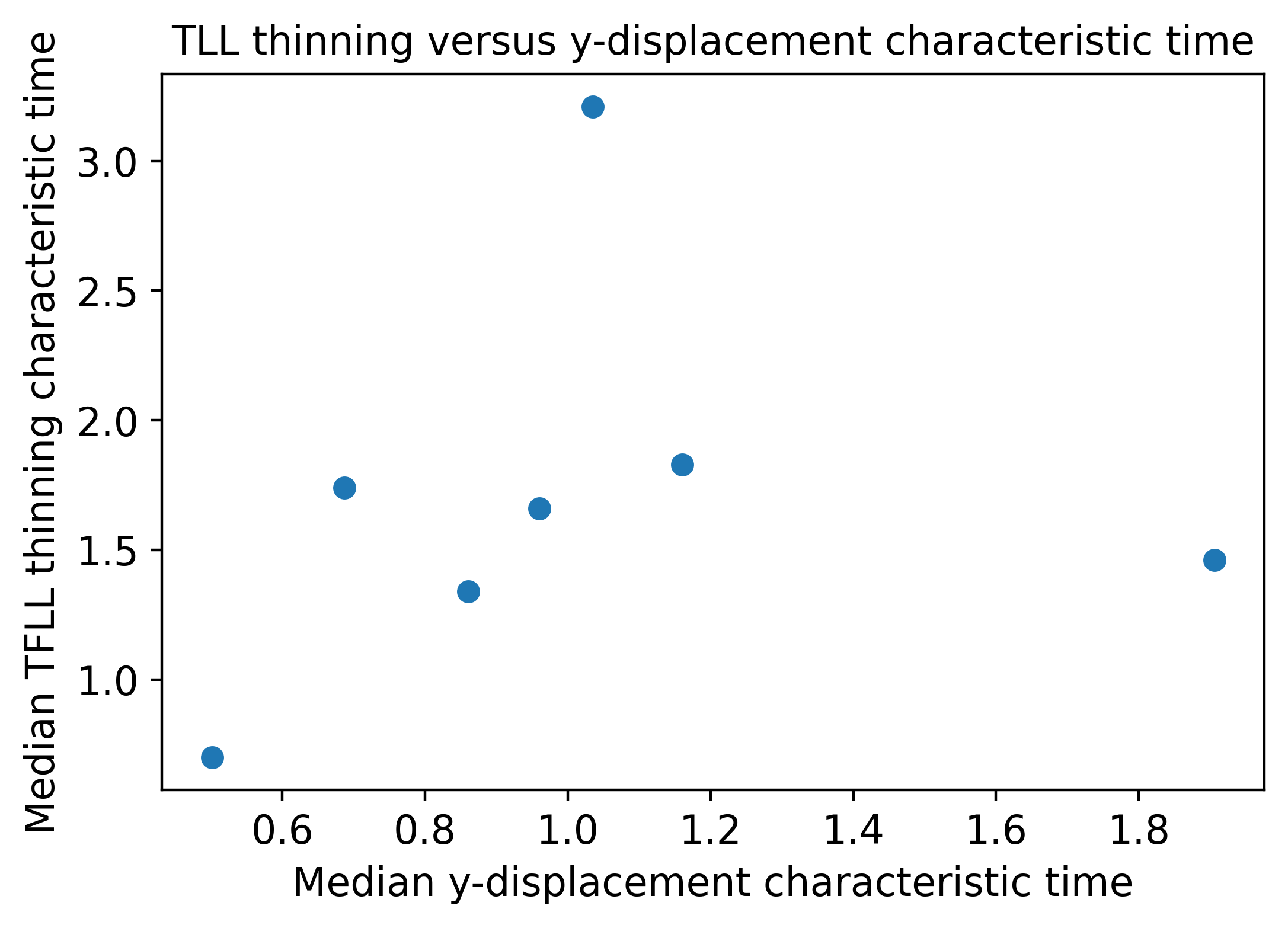}
      \caption{}
      \label{fig:10}
    \end{subfigure}\hfil 
    \caption{\Cref{fig:9} shows that there is a positive correlation between the OSDI score and y-displacement characteristic time. Similarly, \Cref{fig:10} demonstrates the relationship between the TFLL thinning and y-displacement characteristic times.}
    \label{impact}

\end{figure}

\begin{table}[H]
\tymax=300pt
  
  \centering
  \begin{tabulary}{\linewidth}{@{}JLCCC@{}}
    
    \toprule

    Computed y displacement $\lambda$ & Annotation y displacement $\lambda$ & Computed x displacement $\lambda$ & Annotation x displacement $\lambda$\\
    \midrule
    1.05 & 1.17 & 5.18 & 2.32 \\
    1.37 & 1.90 & 1.22 & 1.44  \\
    0.41 & 0.41 & 0.37 & 0.81 \\
    0.51 & 0.46 & 1.49 & 1.60 \\
    0.49 & 0.47 & 7.04 & 7.27 \\
   
    \bottomrule
  \end{tabulary}

  \caption{\label{compvsann}This table reports computed characteristic times from the tracking method and the associated annotation characteristic times for both y and x displacement.}
\end{table}

\section{Results and Discussion}
In order to validate the method, we have selected 11 videos. All of the inter-blink periods are tracked with the described methodology and the x and y displacements are each fit with an exponential decay curve: $\rho\cdot \exp(- \sfrac{t}{\lambda}) + c$. $\lambda$ is the characteristic time. We find that Farneback's dense optical flow algorithm applied on the Laplacian pyramid image performs the best for tracking the lipid layer spread. \Cref{compvsann} compares the characteristic times from the specified tracking method against the ground truth annotations for the five randomly selected inter-blinks and demonstrates that with further refinement, this method holds significant promise for tracking tear lipid layer dynamics. Furthermore, our work confirms that the change in velocity decreases in proportion to the rate of tear film thinning as shown in \Cref{fig:10}. Finally, \Cref{fig:9} shows that there is a positive correlation between the Ocular Surface Disease Index(OSDI) score and the y-displacement characteristic time. This confirms that a numerical analysis of the tear film lipid layer spread corresponding to other validated DED instruments is feasible. A refined tool based on this methodology holds significant promise for future research and clinical care in DED. 

\section{Acknowledgements}
This work was supported by R21EY033881 (Lin/Yu), UCB-CRC Research Fund 51194 (Lin) and Roberta Smith Research Fund 13681 (Lin).



\printbibliography

@misc{easytearwebsite,
%     title     = "EASYTEAR",
%     url       = "https://www.easytear.it/en/"
% }

@article{craig2017tfos,
  title={TFOS DEWS II definition and classification report},
  author={Craig, Jennifer P and Nichols, Kelly K and Akpek, Esen K and Caffery, Barbara and Dua, Harminder S and Joo, Choun-Ki and Liu, Zuguo and Nelson, J Daniel and Nichols, Jason J and Tsubota, Kazuo and others},
  journal={The ocular surface},
  volume={15},
  number={3},
  pages={276--283},
  year={2017},
  publisher={Elsevier}
}

@article{yokoi2008rheology,
  title={Rheology of tear film lipid layer spread in normal and aqueous tear--deficient dry eyes},
  author={Yokoi, Norihiko and Yamada, Hideaki and Mizukusa, Yutaka and Bron, Anthony J and Tiffany, John M and Kato, Takahisa and Kinoshita, Shigeru},
  journal={Investigative ophthalmology \& visual science},
  volume={49},
  number={12},
  pages={5319--5324},
  year={2008},
  publisher={The Association for Research in Vision and Ophthalmology}
}

@article{kass1988snakes,
  title={Snakes: Active contour models},
  author={Kass, Michael and Witkin, Andrew and Terzopoulos, Demetri},
  journal={International journal of computer vision},
  volume={1},
  number={4},
  pages={321--331},
  year={1988},
  publisher={Springer}
}

@article{rosten2008faster,
  title={Faster and better: A machine learning approach to corner detection},
  author={Rosten, Edward and Porter, Reid and Drummond, Tom},
  journal={IEEE transactions on pattern analysis and machine intelligence},
  volume={32},
  number={1},
  pages={105--119},
  year={2008},
  publisher={IEEE}
}

@book{lucas1981iterative,
  title={An iterative image registration technique with an application to stereo vision},
  author={Lucas, Bruce D and Kanade, Takeo and others},
  volume={81},
  year={1981},
  publisher={Vancouver}
}

@inproceedings{farneback2003two,
  title={Two-frame motion estimation based on polynomial expansion},
  author={Farneb{\"a}ck, Gunnar},
  booktitle={Scandinavian conference on Image analysis},
  pages={363--370},
  year={2003},
  organization={Springer}
}

@article{kowalski2021hybrid,
  title={Hybrid FPGA-CPU pupil tracker},
  author={Kowalski, Bartlomiej and Huang, Xiaojing and Steven, Samuel and Dubra, Alfredo},
  journal={Biomedical Optics Express},
  volume={12},
  number={10},
  pages={6496--6513},
  year={2021},
  publisher={Optical Society of America}
}

@article{farrand2017prevalence,
  title={Prevalence of diagnosed dry eye disease in the United States among adults aged 18 years and older},
  author={Farrand, Kimberly F and Fridman, Moshe and Stillman, Ipek {\"O}zer and Schaumberg, Debra A},
  journal={American journal of ophthalmology},
  volume={182},
  pages={90--98},
  year={2017},
  publisher={Elsevier}
}

\end{document}